%% file: 0_main_file_aacl.tex
\pdfoutput=1

\documentclass[11pt]{article}

\usepackage[]{acl}

\usepackage{times}
\usepackage{latexsym}

\usepackage[T1]{fontenc}
\usepackage{titlesec}
\usepackage[utf8]{inputenc}

\usepackage{microtype}
\usepackage{graphicx}
\usepackage{multirow}
\usepackage{booktabs}
\usepackage{enumitem}
\usepackage{arydshln}
\usepackage{algorithm}
\usepackage{algpseudocode}
\usepackage{amsmath}
\usepackage[multiple]{footmisc}
\usepackage{amsfonts}

\mathchardef\mhyphen="2D 

\title{Multilingual CheckList: Generation and Evaluation}

\author{Karthikeyan K\textsuperscript{3}*, Shaily Bhatt\textsuperscript{1}*, Pankaj Singh\textsuperscript{1}, Somak Aditya\textsuperscript{4},\\\bf{Sandipan Dandapat\textsuperscript{2}, Sunayana Sitaram\textsuperscript{1}, Monojit Choudhury\textsuperscript{1}}
\\
\textsuperscript{1} Microsoft Research, Bengaluru, India
\\
\textsuperscript{2} Microsoft R\&D, Hyderabad, India
\\
\textsuperscript{3} Department of Computer Science, Duke University
\\
\textsuperscript{4} Department of CSE, IIT Kharagpur
\\
\texttt{karthikeyan.k@duke.edu, saditya@cse.iitkgp.ac.in,}
\\
\texttt{\{t-shbhatt,t-pasingh,sadandap,sunayana.sitaram,monojitc\}@microsoft.com}
        }

\begin{document}
	\maketitle

	\begin{abstract}
	        Multilingual evaluation benchmarks usually contain limited high-resource languages and do not test models for specific linguistic capabilities.
    		CheckList \citep{checklist-paper} is a template-based evaluation approach that tests models for specific capabilities.
    		The CheckList template creation process requires native speakers, posing a challenge in scaling to hundreds of languages. In this work, we explore multiple approaches to generate Multilingual CheckLists. We device an algorithm -- \textbf{T}emplate \textbf{E}xtraction \textbf{A}lgorithm (TEA) for automatically extracting target language CheckList templates from machine translated instances of a source language templates. We compare the TEA CheckLists with CheckLists created with different levels of human intervention. We further introduce metrics along the dimensions of \textit{cost}, \textit{diversity}, \textit{utility}, and \textit{correctness} to compare the CheckLists. We thoroughly analyze different approaches to creating CheckLists in Hindi. Furthermore, we experiment with 9 more different languages. 
    		We find that TEA followed by human verification is ideal for scaling Checklist-based evaluation to multiple languages while TEA gives a good estimates of model performance. We release the code of TEA and the CheckLists created at \href{https://aka.ms/multilingualchecklist}{aka.ms/multilingualchecklist}
	\end{abstract}




\section{Introduction}
\input{aacl/1_intro}

\section{TEA: Template Extraction Algorithm}
\input{aacl/2_background}
\input{aacl/3_TEA}

\section{Multilingual CheckList Generation}
\input{aacl/4_multilingual_checklist_generation}

\section{Evaluation Metrics}
\input{aacl/5_metrics}

\section{Hindi CheckLists and Results}
\input{aacl/6_hindi_checklists}

\section{CheckLists in Multiple Languages}
\input{aacl/7_multilingual_experiments}

\section{Limitations}
\input{aacl/8_discussion}

\section{Conclusion}
\input{aacl/9_conclusion}

\paragraph{Acknowledgements} We thank the annotators and their team who facilitated the user studies in the paper. We are grateful to our colleagues at MSRI for being the pilot users. We thank Kalika Bali, Balakrishnan Santhanam, and Prasenjit Rath for their thoughtful guidance throughout the work. We thank Gauri Kholkar for her engineering assistance.

\bibliography{custom}
\bibliographystyle{acl_natbib}

\appendix

\input{aacl/app_1_tea_details}

\input{aacl/app_2_checklist_capabilities_tested.tex}

\end{document}

%% file: aacl/1_intro.tex
Multilingual transformer based models \citep{bert,xlmr,liu2020multilingual,xue2021mt5} have demonstrated commendable zero \& few-shot capabilities. Their performance is typically evaluated on benchmarks like XNLI \citep{xnli}, XGLUE \citep{xglue}, XTREME \citep{hu2020xtreme} \& XTREME-R \cite{ruder2021xtremer}. However, this evaluation paradigm has a number of limitations including: First, most of these datasets are limited to a few high resource languages~\citep{pmlr-v119-hu20b, wang-etal-2020-extending, vulic-etal-2020-multi}, except for a few tasks (e.g., NER, POS \cite{ahuja-etal-2022-beyond, bhatt-2021-on}). Second, creating high quality test sets of substantial size for many tasks and languages is prohibitively expensive. Third, state-of-art models are known to learn spurious patterns to achieve high accuracies, saturating performance on these test-benches, yet performing poorly on often much simpler real world cases~\citep{balanced_vqa_v2, gururangan-etal-2018-annotation, glockner2018breaking, tsuchiya2018performance, geva2019we}. Fourth, these benchmarks do not evaluate models for language specific nuances \cite{checklist-paper}. Lastly, this evaluation approach does not provide any insights into where the model is failing \cite{wu-etal-2019-errudite}. These limitations lead to the need of interactive, challenging, and much larger testing datasets (like \cite{big-bench, dyna-bench}) and more holistic approaches to evaluation (like \citet{checklist-paper}).

CheckList \cite{checklist-paper} is an evaluation paradigm that systematically tests the various {\em(linguistic) capabilities} required to solve a task. It allows creation of large and targeted test sets easily using various abstractions. 
Specifically, users can generate {\em templates}, essentially sentences with {\em slots} that can be filled in with a dictionary of {\em lexicons} to generate test {\em instances}. CheckList templates are created by native speakers. \citet{ruder2021xtremer} introduce Multilingual Checklists created by human translation from English CheckList for 50 languages for a subset of tests on Question Answering. However, since CheckLists are task \& language specific, human creation or translation of CheckLists remains extremely resource-intensive. 

In this paper, we introduce an automatic approach to creating Multilingual CheckLists. We devise the  \textbf{T}emplate \textbf{E}xtraction \textbf{A}lgorithm (TEA) for extracting templates in a {\em target} language from the translated instances of a {\em source} language CheckList (here English) automatically (\S\ref{3_tea}). We also experiment with semi-automatic and manual approaches for Multilingual CheckList creation (\S\ref{4_multilingual_checklist_generation}). In the semi-automatic approach (TEA-ver), we ask human annotators to verify and correct the templates created by TEA. In the manual approach, we ask annotators to create CheckLists in two ways: first, by translation of English CheckList to the target language (t9n) (same as \citet{ruder2021xtremer}); Second, by giving a description of the task and capabilities to create CheckLists from scratch (SCR) (same as original English CheckLists creation \cite{checklist-paper}).Using these four approaches, we create CheckLists for Sentiment Analysis (SA) and Natural Language Inference (NLI) in Hindi (\S\ref{6_hindi_checklists}). We demonstrate broad applicability of TEA by generating CheckLists in additional 9 typologically diverse languages (Gujarati, French, Swahili, Arabic, German, Spanish, Russian, Vietnamese, Japanese) and TEA-ver CheckLists in 3 of them (\S\ref{7_multilingual_experiments}).

Evaluation of CheckLists is non-trivial. For thorough comparisons, we propose evaluation metrics along four axes: {\em utility}, {\em diversity}, {\em cost} \& {\em correctness} (\S\ref{5_metrics}). 
Our evaluation indicates that CheckLists created using TEA are not only cost-effective but also useful and diverse, with comparable quality to the manually and semi-automatically created CheckLists. Experiments on typologically diverse languages show that TEA CheckLists provide a good estimate of the failures of the model, and thus can be used even in the absence of resources to verify them or create human-annotated gold test-sets.

To summarize, our contributions are:
a) We propose TEA (\textbf{T}emplate \textbf{E}xtraction \textbf{A}lgorithm) to extract templates in a target language using translated instances of a source CheckList.
b) We  experiment with varying degrees of human intervention, comparing semi-automatic \& manual approaches of Multilingual CheckList creation with TEA, to understand the best utilization of the human effort. 
c) We introduce evaluation metrics along the axes of \textit{utility}, \textit{diversity}, \textit{cost}, and \textit{correctness} for in-depth comparison of the the CheckLists. 
d) We will release all the 4 CheckLists in Hindi for SA and NLI, TEA CheckLists in 9  languages for SA and TEA-ver CheckLists in 3 languages for SA.

We release the code of TEA and the CheckLists created at \href{https://aka.ms/multilingualchecklist}{aka.ms/multilingualchecklist}


%% file: aacl/2_background.tex
Terminology (consistent with \citet{checklist-paper}):

\noindent
\textbf{\textit{Linguistic} capabilities:} These are capabilities tested for a particular task. For e.g, negation.

\noindent
\textbf{Templates:} These are sentences with slots. For e.g, `\{CITY\} is beautiful'. Here, `\{CITY\}' is a slot. Templates can have any number of slots.

\noindent
\textbf{Lexicon keys and values:} This a dictionary of values. In the above example, `CITY' is the key. Values are the words that would be filled in the slots (replacing the keys) like `New Delhi', `New York', `London', etc. We use the notation `CITY = [`New Delhi', `London', `New York'] ' for lexicons.

\noindent
\textbf{Instances:} These are test sentences created by inserting lexicon values in templates
. In the above example, the instances formed are: `New York is beautiful', `London is beautiful', etc.

%% file: aacl/3_TEA.tex
\label{3_tea}

The CheckList paradigm allows creation of large number of test instances. For multilingual evaluation, these can then be translated to the target languages using Machine Translation. However, there are limitations to this approach. Firstly, a large machine translated test set is difficult to be verified by humans, as one would have to go through every example. Second, it defeats the purpose of abstraction that CheckLists facilitates. And third, the quality of this test set will be directly impacted by the quality of the MT system. This results in the need to generate templates in the target language so that these can be utilized and verified in the same fashion as the template sets in the source language.

Our early experiments suggested that due to word order and syntactic differences between languages, both: 1) a word-to-word or heuristic translation of the template
and 2) extraction of template from a single source instance (such as by simply replacing one word with other in a single translated instance)
do not work well for template translation.
This necessities a non-trivial algorithm that can extract templates given a set of instances.

We propose the \textbf{T}emplate \textbf{E}xtraction \textbf{A}lgorithm or TEA, to automatically extract template sets given an input a set of instances. In this paper, these input instances for TEA are obtained by machine translating instances created from the source CheckList template sets. We use machine translation to reduce cost and human effort, but the algorithm can be used with any input set of instances, i.e it would work with human-translated instances.

\begin{figure*}[ht]
\centering
\includegraphics[scale=0.18]{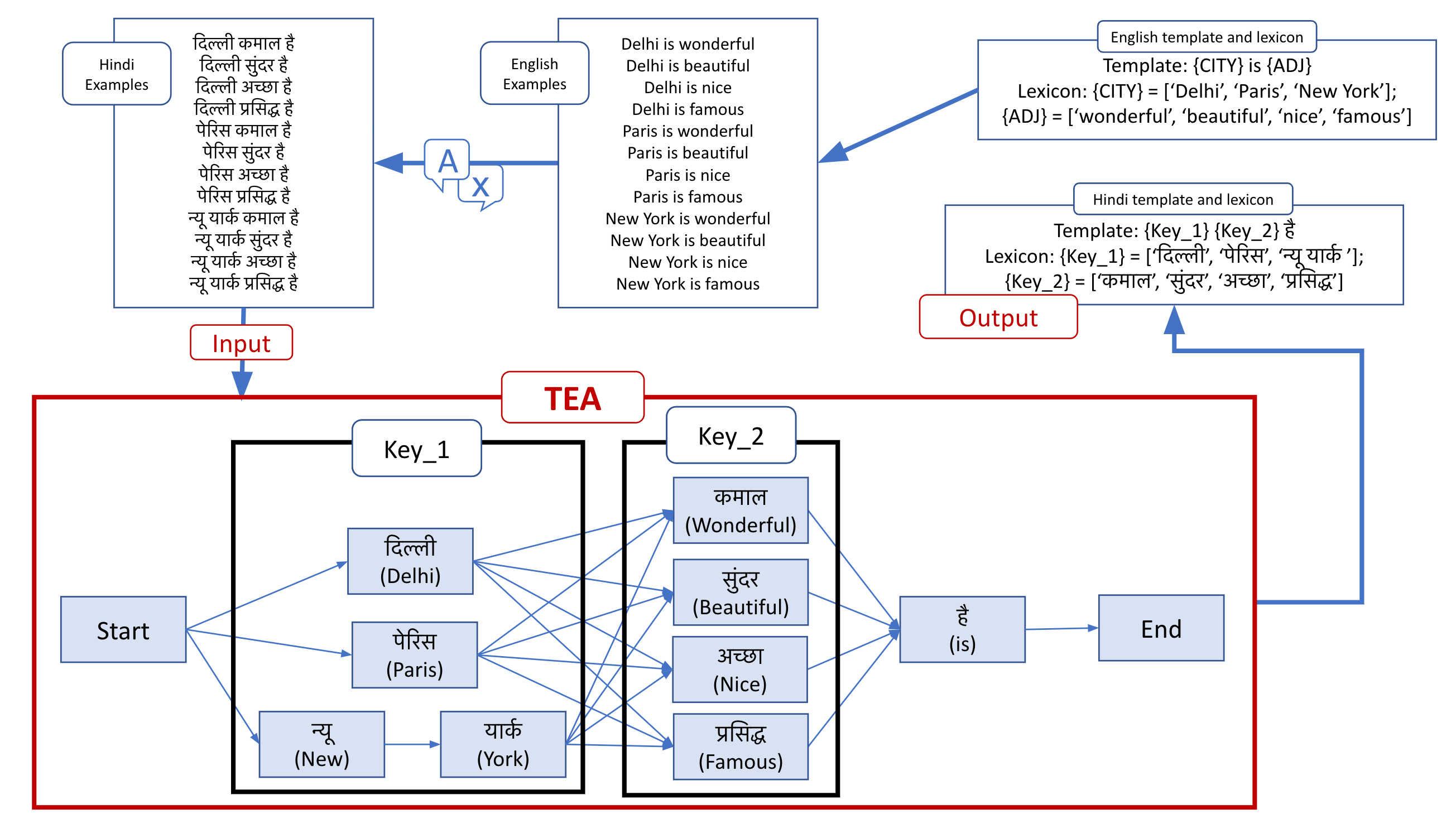}
\caption{TEA treats sentences as a directed acyclic graph \& recursively replaces lexicon values with keys.}\label{fig:tea}
\end{figure*}

Briefly, TEA is a recursive approach to extract templates from input instances by treating every input instance as directed acyclic graph of the words. TEA combines the instances with similar structure into a single template by recursively merging instances and replacing terminals (or lexicon values) with non-terminals (or lexicon keys).

Figure \ref{fig:tea} shows how TEA creates lexicon keys by combining instances from the translated instances. We assume English (EN) to be the source language and Hindi (HI) to be the target language. The pipeline starts with an EN template, the instances are created by replacing the lexicon values in the templates, that are then machine translated to get HI instances. These instances function as input to TEA which then recursively groups instances using non-terminals to form templates. The entire process of template extraction is repeated for every EN template, resulting in the HI template set. The TEA has 3 steps which we describe as follows (pseudocode and details are in appendix \ref{app:tea_details}):

\paragraph{Step 1: Grouping Terminals into Non-Terminals: } First, we convert the Hi instances into a directed acyclic graph whose nodes are unique words (or tokens). There is an edge from node A to B if word B follows word A in at least one of the input instances. In this directed graph (see Fig.~\ref{fig:tea}), between any two nodes, if there are multiple paths of length less than equal to $k+1$ (we set $k$ to 2), we concatenate the intermediate words in the path (with space in between them) and treat them as terminals. This set of terminals, between the two nodes, are grouped together represented by a non-terminal symbol (for example Key\_1 and Key\_2 in Fig.~\ref{fig:tea}). This step corresponds to lexicon formation; the non-terminal extracted here are essentially keys of the lexicon \& the terminals constituting them are the lexicon values for the slots of a template.

\paragraph{Step 2: Template Extraction} Using a set of Hi instances, $S = \{s_1, s_2, \ldots s_N \}$, and all non-terminals $v_i = [ {w_{i1}, w_{i2}, \ldots}]$, where $w_{ij}$ are terminals (obtained step 1), TEA outputs a set of templates $\hat{T} = \{t_1, t_2, ...\} $ such that $\hat{T}$ can generate all the examples in $S$ using the given non-terminal and their corresponding terminals. For each sentence $s_i$, we generate a set of candidate templates, $T_i = \{t_{i1}, t_{i2}, ...\}$, such that $s_i$ belongs to the set of examples generated by each $t_{ij}$. To find the minimal template set, i.e $\hat{T}$ that covers all examples is treated as a set cover problem and we use a greedy approximation to find this set.

\paragraph{Step 3: Combine Steps 1 and 2} The above template extraction process, while resulting in correct outputs, may be computationally expensive due to translation noise\footnote{The translated sentences may not fit into 1 template. Or, the algorithm may produce a set of distinct non-terminals with common or overlapping terminals. For e.g, we may get two non-terminals with their corresponding terminals such as ``\{Paris, New York\}''  \&  ``\{London, New York, Delhi\}''.} and its time-complexity which is exponential on the number of non-terminals. To mitigate this, we follow an iterative approach where instead of using all the extracted non-terminals (along with their terminals), we initialize the set of non-terminals with an empty set and iteratively add the most useful non-terminals (with their corresponding terminals) to this set. 

Note that, TEA can generate multiple templates for the set of instances (all of which might be generated from a single source template). This design is intentional and desirable as due to morpho-syntactic complexities (e.g, grammatical gender), it is likely that all instances in a target language will not fit into a single template.



%% file: aacl/4_multilingual_checklist_generation.tex
\label{4_multilingual_checklist_generation}


We now describe the various ways in which multilingual checklists can be generated, ranging from fully automatic to fully manual approaches.

\paragraph{Using TEA} We start with a source language (En) CheckList template and generate instances by replacing lexicon values in templates. These instances are translated using an MT system. The translated instances now serve as the input to the TEA and target language (HI) CheckList template is extracted. The process (Fig. \ref{fig:tea}) is repeated for all En templates to form the complete HI CheckList.

\paragraph{TEA with Verification (TEA-ver)} This is a semi-automatic approach, where we ask a human-{\em verifier} to verify and correct the CheckLists generated using TEA. The verifiers (or annotators) are provided with a set of templates and lexicons generated using the TEA pipeline, along with the original source langauge CheckList and description of the capabilities. The annotators are instructed to {\em verify} the target language templates for (grammatical) correctness. They can delete or edit the incorrect templates. They can also add any missing templates that they think are significantly important (cover too many missed instances).

\paragraph{Translating source CheckList (t9n)} This is a completely manual approach, but relies on a source language (here, En) CheckList. The annotators are provided with the En templates, lexicons and the descriptions of capabilities. They are tasked to translate the templates and lexicons into the target language. If a source template cannot be translated to a single target template (such as due to divergent grammatical agreement patterns),  annotators are instructed to include as many variants as necessary. This approach is same as that used by \citet{ruder2021xtremer} to create multilingual CheckList.

\paragraph{Generating CheckList from scratch (SCR)} This is a completely manual approach of creating CheckLists from scratch, not relying on any source CheckList. Here, the CheckList templates are generated in the same manner as generated in by humans in \citet{checklist-paper}. That is, human annotators are provided with a description of the task and capabilities and are instructed to develop the templates and lexicon, directly in the target language. In our pilot we found that users were better able to understand the capabilities with some examples as opposed to only from the description, so we also provided them with a couple of examples, in English, for each capability. 


%% file: aacl/5_metrics.tex
\label{5_metrics}

Comparison of CheckLists is non-trivial. Firstly, CheckLists cannot be evaluated using absolute metrics, comparisons can only be relative \cite{humaneval_eacl_case_study}. Further, the question of what constitutes a better CheckList can be answered in multiple ways. For example, if a CheckList A can help discover (and/or fix) more bugs than CheckList B, CheckList A could be more useful. On the other hand, variability of instances may be desirable. If CheckList B generates more diverse instances as compared CheckList A, even though it discovers less bugs, B could be considered better as it allows testing of the system on a broader variety of instances. Finally, in practical scenarios, cost and correctness are both important factors for generating the CheckList.

We thus propose evaluation metrics along 4 dimensions: 
a) {\em utility} for discovery and fixing bugs; 
b) {\em diversity} in the generated instances; 
c) {\em cost} of generating templates. 
d) {\em correctness} of templates. 


\subsection{Utility}

\paragraph{Failure Rate (FR)} Here, we measure the percentage of instances generated by the CheckList that the model failed on averaged over all the capabilities.\footnote{Unless mentioned otherwise, we report macro-averages across capabilities.} The numbers are reported for XLM-R fine-tuned with English task data from standard datasets (SST-2 for SA and mNLI for NLI). Effectively, we measure the FR on zero-shot transfer from English to the target language. For FR, the higher the value the better the CheckList.

\paragraph{Augmentation Utility (Aug)} These metrics aims to test the utility of CheckList in fixing failures using data augmentation following \citet{humaneval_eacl_case_study}. This is done in two ways:

\textbf{(a) From Scratch (Aug-0)}: Here, we fine-tune XLM-R directly using CheckList instances.

\textbf{(b) On Fine-tuned model (Aug-CFT)}: Here, XLM-R is first fine-tuned with English task data (SST-2 for SA and mNLI for NLI) and then further continually fine-tuned using CheckList instances.

In both cases, we first generate all instances using the CheckLists being compared. We retain a maximum of 10k instances per capability for each CheckList. The instances are then randomly split into train and test sets in 70:30 ratio. The training data (of the corresponding CheckList) is used for the augmentation as described above. The test sets, generated from all the CheckLists being compared are combined together to form a common test set and accuracy on this set is reported. Intuitively, this aims to determine the utility of the CheckList's instances for fixing failures using augmentation. For both the Augmentation metrics, higher is better.

\subsection{Diversity}

\paragraph{Number of templates (\#temp) and lexicon values (\#lexv)} The simplest way to measure the diversity is the number of distinct templates and lexicon values (or terminals). Higher number of templates and lexicon values means more diversity. 

\paragraph{Normalized Cross-Template BLEU (CT-BLEU)} To measure the diversity between the templates, we measure the BLEU score (or similarity) for every instance generated by a template 
with the the instances generated by all other templates in the CheckList
. Since this score is sensitive to the number of templates in the Checklist, we normalize the score by the number of templates in the set. Lower CT-BLEU is indicative of better CheckList as it indicates more diverse instances from templates.

\subsection{Cost}

\paragraph{Time per template (TpT)} We define the {\em cost} of creation of these Checklists simply as the human time required. Since different methods or users can create substantially different number of templates per capability, we measure the {\em mean time taken} (TpT) for creation from scratch (SCR), translation (t9n) and verification (TEA-ver) of a {\em template} as the measure of the cost. A better CheckList for practical purposes would have lower TpT.

\subsection{Correctness}

Here, we assume that templates generated with any amount of human intervention (manual or semi-automatic) would always be correct. As a result, we calculate correctness only for TEA templates. We define the correctness of TEA templates with respect to TEA-ver templates. This is because during creation of templates by the TEA-ver process annotators correct or remove templates. Thus, only correct TEA template are left unedited. Therefore, in order to estimate the correctness of the TEA templates, we compute the following two metrics.

\paragraph{Failure Rate Difference (FR-diff)} It is possible that the model fails in some cases if the input instance is not well-formed. As a result, the difference between the failure rates induced by TEA-ver templates (which always lead to well-formed instances) and that of TEA templates (which could lead to some ungrammatical instances) will give an estimate to the correctness of TEA templates. As a result, we define this metric as simply the difference between the FR of TEA and TEA-ver.

\paragraph{Precision and Recall (P/R)} Since during the TEA-ver process, annotators edit or remove incorrect templates, only the correct templates that were generated by TEA are left as is. Therefore, in order to estimate the correctness of the TEA, we compute the precision and recall of the TEA template set, with respect to TEA-ver template set. We define 
match when the templates are same and the lexicon values of either one is a subset of the other, implying they will generate similar set of examples.

%% file: aacl/6_hindi_checklists.tex
\label{6_hindi_checklists}

We start with Hindi (Hi) as the target language, create CheckLists using all 4 methods from \S\ref{4_multilingual_checklist_generation} and evaluate them using the metrics from \S\ref{5_metrics}. Hi has significant syntactic divergence from the source language (here English (En)) and uses a different script. Hi is a mid-resource language with reasonably good publicly available En-Hi MT systems. We argue that if TEA works well in the En-Hi pipeline, it would also work for most other high to mid resource languages with reasonable MT systems and similar or less syntactic divergence from En, which we also substantiate by performing additional multilingual experiments in \S\ref{7_multilingual_experiments}. 

\subsection{Experiment Design}

We create and evaluate Hi CheckLists for 2 tasks, Sentiment Analysis (SA) and Natural Language Inference (NLI). For SA, we choose $5$ capabilities  namely Vocabulary, Negation, Temporal, Semantic Role Labeling and Relational, and their associated Minimum Functionality Test (MFT) templates from~\citet{checklist-paper} as our source CheckList. For NLI, we choose co-reference resolution, spatial, conditional, comparative and causal reasoning as capabilities and their associated templates from~\citet{nlichecklist}. We refer readers to Appendix \ref{app:capabilities} for details about these capabilities.

Following \citet{checklist-paper}, we chose $6$ software developers as our {\em annotators}, who are knowledgeable in NLP. All users are native speakers of Hi and have near-native En fluency.\footnote{Educated for 15+ years in English} We expect developers to be the actual users of the approach, as it is usually a developer's job to find and fix bugs. The annotators
were given a detailed description of expectations along with examples (both in En and Hi). Furthermore, during our pilot study, we found some of the common errors users make, and to mitigate those we provided a list of common errors illustrated with simple examples.  

Each of the $6$ annotators was randomly assigned a CheckList creation approach that requires human intervention. Thus, we had 2 annotators each for the SCR, t9n and TEA-Ver setups. They carry out the process independently for both SA and NLI. The same description of capabilities and examples are used for all the experimental setups. Similarly, the same source templates and lexicons are used for t9n, TEA-ver and TEA. For the TEA pipeline, we used Bing Translator API for translating En instances to Hi. While reporting the results, we report the average metrics of both annotators.


\subsection{Results}

	\begin{table*}[!ht]
        \small
		\centering
		\begin{tabular}{ll | llll | llll}
			\toprule
			\textbf{Metric} & & \multicolumn{4}{|c|}{\textbf{Sentiment Analysis}} & \multicolumn{4}{c}{\textbf{NLI}} \\
			\midrule 
			& & \textbf{SCR} & \textbf{t9n } & \textbf{TEA-ver}  & \textbf{TEA} & \textbf{SCR} & \textbf{t9n } & \textbf{TEA-ver}  & \textbf{TEA} \\
			\midrule
			\textbf{Utility} &  \textbf{FR} & 6.7 & 16.5 & \textbf{19.7} & 19.3 & \textbf{60.3} & 53.4 & 45.1 & 48.4 \\
			&  \textbf{Aug-0} & 49 & 52.4 & 50.6 & \textbf{67.1} & 16.2 & 50.9 & \textbf{58.4} & 52.5 \\
			 &  \textbf{Aug-CFT} & 86.8 &  89.9 &  \textbf{95.3} & \textbf{95.3} & 70.1 & 81.2 & 79.4 & \textbf{83.2} \\
		
			\midrule
			\textbf{Diversity}&  \textbf{\#~temp} & 17 & 44.5 & 86.0 & \textbf{105} & 16 & 22.5 & 51.5 & \textbf{54} \\
			 & \textbf{\#~lexv} &  35.0 & 41.5 & 109 & \textbf{147.0} & 38.5 & 56.5 & 88.5 & \textbf{98.0} \\ 
			& \textbf{CT-BLEU*} & 0.511 & 0.142 & 0.096 & \textbf{0.087} & 0.564 &  0.307 & 0.216 & \textbf{0.169}\\
			\midrule
			\textbf{Cost} &   \textbf{TpT* (mins)} & 5.38 & 2.07 & 1.77 & \textbf{0} & 4.69 & 3.67 & 1.91 & \textbf{0} \\
   \midrule
   \textbf{Correctness} &   \textbf{FR-diff*} & - & - & - & 0.4 & - & - & - & 3.3 \\
    &   \textbf{P/R} & - & - & - & 0.64/0.61 & - & - & - & 0.67/0.63 \\
			\bottomrule
		\end{tabular}
		\caption{\label{tab:metrics} Comparison of the 4 approaches across two tasks for Hindi. *Lower is better; for rest higher is better.}
		
	\end{table*}
	
Table \ref{tab:metrics} reports the metrics (\S\ref{5_metrics}) for the 4 methods.
	
The trends for {\em cost} or TpT are consistent with expectations. Creating CheckLists from Scratch (SCR) takes the most time, as the user has to think and create the templates. t9n requires manual translation and is quicker than SCR but slower than TEA-ver, which just requires verification and correction on templates generated by TEA. We do not factor in the time required to create the source En Checklist, because 1) It is common to all of these 4 approaches and sourced from existing literature; and 2) it is a one-time effort which can be reused for generation of CheckLists in many target languages, leading to a very low amortized cost.
	
In {\em diversity} metrics TEA generates the most diverse templates, closely followed by TEA-Ver. t9n is much less diverse, and SCR has the least diversity. We found that, the users created very few templates for SCR, perhaps because it is difficult to decide what would be a good number of templates. We also observe that TEA generates a largest number of templates. The source checklists had 32 (74) and 18 (76) templates (lexicon values) for SA and NLI, respectively. Thus on average, a source template generates around 3 target templates, which is primarily due to syntactic divergence between the En and Hi. These numbers are reduced in TEA-ver, most likely because not all of the TEA templates are perfect and human annotators merge or delete some of them during the verification.

The trends in {\em utility} metrics are varied. In SA, TEA-ver templates induce highest FR and TEA is a close second. However, for NLI, SCR CheckList induces the highest failure, followed by t9n. This might be due to the task complexity. We leave further exploration on the co-relation of task complexity and efficacy of TEA to future work. TEA has the highest Aug-0 and Aug-CFT values except one case where it is a close second, indicating that the instances generated by TEA CheckLists are effective in fixing failure by augmentation. TEA-ver has values that close to TEA for these metrics\footnote{TEA and TEA-ver have a substantial overlap, and thus, augmentation of one typically helps with the other. This explains the high AUG-0 and AUG-CFT values for these setups.}. 
	
In terms of {\em correctness}, based on P/R of TEA with respect to TEA-ver, we find that that around a third of the TEA templates had to be significantly edited or removed. Despite this, from FR-diff, we see that the FR generated by TEA is fairly close to the FR generated by TEA-ver. Additionally, even the numbers of other utility metrics are also comparable. This indicates that even the unverified templates (from TEA) which may generate some ungrammatical instances, can give very close estimates of the failure rates and augmentation accuracy to human-verified template sets. This is a positive finding, because while TEA-ver is more reliable, but when resources to get TEA templates verified are not available, despite imperfections, TEA CheckLists can be used for evaluation. 

Finally, we would like to point out some of the qualitative differences that we saw in the CheckLists created by these different methods which are hard to articulate through metrics. In particular, we saw that CheckLists created from scratch tend to capture cultural context better. For example, annotators use Indian names in the lexicon values as opposed to western names that get generated due to translations in all other 3 approaches. However, while difficult for TEA, this entity recontextualization is fairly easy for the other two approaches where humans are involved. We also find that the template sets of TEA-ver and t9n are overlapping. This is because of the setup, where t9n is directly translated at template level and TEA-ver is obtained after correcting the templates obtained from translated instances. The major difference occurs in the amount of time taken as correcting templates is faster than translating them.
	
Thus overall, we conclude that TEA followed by human verification, or TEA-ver would be an ideal approach for scaling CheckList evaluation to multiple languages. That said, the fully automatic TEA approach is even more cost-effective and almost equally reliable to the TEA-ver approach, making it suitable for large-scale multilingual CheckList generation with extremely limited resources.

%% file: aacl/7_multilingual_experiments.tex
\label{7_multilingual_experiments}

\begin{table*}[ht]
\centering
\small
\resizebox{\textwidth}{!}{%
\begin{tabular}{llllllll} 
			\toprule
			\textbf{Language}     &               & \textbf{Vocabulary} & \textbf{Temporal} & \textbf{Fairness} & \textbf{Negation} & \textbf{SRL} & \textbf{Robustness}  \\ 
			\midrule
			\textbf{English} & FR (SCR)    & 24.21             & 1.8               & 94.35             & 48.16             & 35.94        & 42.58                \\ 
			\midrule
			\midrule
			\textbf{Gujarati} & FR (TEA)   & 39.12   & 34.97    & 87.46     & 51.84  & 47.37 & 52.09, 51.54 \\
                            & FR (TEA-ver) & 29.09 & 32.18 & 88.72 & 55.15 & 46.8 & 51.54 \\
                            & FR-diff   & 10.09   & 2.79   & 1.26   & 3.3    & 0.57  & 0.55    \\
   
			\midrule
			\textbf{French} & FR (TEA)  & 20.27 & 11.22  & 86.52 & 56.55 & 40.09  & 46.77   \\
                & FR (TEA-ver)  &  21.78   &  11.53    &  86.52   &  61.25 &  40.09   &  47.8    \\
                & FR-diff & 1.51 & 0.31 & 0 & 4.7 & 0 & 1.3\\
			\midrule
			\textbf{Swahili} & FR (TEA)   & 46.04  & 37.5 & 88.86  & 73.32 & 51.87 & 58.45    \\
                            & FR (TEA-ver)   & 38.53  & 43.72 & 90.37 & 73.25 & 46.51  & 55.38   \\
                            & FR-diff & 8.24 & 6.22 & 1.51 & 0.07 & 5.36 & 3.07  \\
			\midrule
			\textbf{Arabic}  & FR (TEA)  & 46.77               & 14.37              & 91.98            & 52.08             & 39.4        & 53.32                \\
			
			\midrule
			\textbf{German}  & FR (TEA)    & 38.45               & 15.59              &85.25            & 47.56             & 43.03        & 44.04                \\
			\midrule
			\textbf{Spanish} & FR (TEA)    & 29.44               & 3.18              &89.45            & 59.41             & 41.39        & 50.1                \\
			\midrule
			\textbf{Russian} & FR (TEA)    & 40.26               & 5.07              &93.67            & 56.13             & 40.3        & 47.61                \\
			\midrule
			\textbf{Vietnamese} & FR (TEA)    & 23.50               & 21.67              &93.22            & 63.05             & 53.12        & 50.97                \\
			\midrule
			\textbf{Japanese} & FR (TEA)    & 26.9               & 24.22              &93.69            & 50.1             & 50.97        & -                \\
			\bottomrule
\end{tabular}%
}
\caption{Failure rates for 9 more languages across 6 capabilities for sentiment analysis. Failure rates of English are for the original templates created manually by annotators (SCR); For Gujarati, French, and Swahili FR for TEA, TEA-ver and FR-diff is reported, for the rest of languages FR for TEA is reported.}
\label{tab:difflangs}
\end{table*}

So far, we see that TEA is cost-efficient in producing effective Hi CheckLists. We now experiment with 9 more typologically diverse languages -- Arabic, French, German, Gujarati, Japanese, Russian, Spanish, Swahili and Vietnamese to evaluate the efficacy of scaling TEA to may languages. We use TEA to automatically generate CheckLists across these languages from the same set of source templates in English for SA across 6 capabilities: Vocabulary, Temporal, Fairness, Negation, Semantic role labeling (SRL) and Robustness. We use the same source En CheckList from the \citet{checklist-paper} and use Bing Translator in the TEA pipeline to translate En instances to the target language.

In Table \ref{tab:difflangs} we report the FR on XLM-R model fine-tuned with SST-2 data; thus, except for En, all other values are for zero-shot transfer to the respective language. The average FR for AMCG is highest for Swahili (59\%), Vietnamese and Gujarati (around 52\%), and lowest for French (43\%), Spanish and German (around 45\%). For English, average FR is 41\%. These trends are consistent with expectation of performance as English, French, and other European languages are high-resourced while Swahili and Vietnamese are very low-resourced. 

For 3 of the target languages, namely French, Gujarati and Swahili, native speakers verified the generated templates and thus, we also report the FR for TEA-ver.\footnote{These languages were selected based on typological, geographical, resource level diversity and access to native speakers.} We observe that the Pearson (Spearman) correlation between TEA and TEA-ver FR values for French, Gujarati and Swahili are 0.99 (1.0), 0.98 (0.89) and 0.97 (0.94) respectively. Furthermore, the difference between FR (FR-Diff) is also low. This implies, similar to our observations from section \ref{6_hindi_checklists}, that one can obtain an extremely accurate assessment of the capabilities of multilingual models just from TEA CheckLists even for low resource languages like Swahili. This re-affirms that despite noise, TEA is able to generate CheckLists that are useful without any human supervision.



%% file: aacl/8_discussion.tex
\label{8_discussion}

In this paper, we introduced the TEA  to generate target language CheckList (templates + lexicon) from the translated instances of source language CheckList. We show that with drastically reduced human effort required for creating CheckList in a new language, the TEA CheckLists provide an accurate estimate of the models' capabilities. However, some of the generated templates/lexicons are noisy and were removed or edited by humans through the TEA-ver process. In this section, we summarize the limitations, common error patterns and suggest some possible ways to resolve them.

\textbf{Agnostic to Semantics} TEA is agnostic of the semantics of the lexicon keys. So, when faced with a set of sentences: \textit{Las Vegas is good.}, \textit{New York is good.}, \textit{New Delhi is good.} and \textit{Las Palmas is good.}, it is unclear whether it should design 1 template \texttt{CITY is good.} with lexicon \texttt{CITY=}\{\textit{Las Vegas, New York, New Delhi, Las Palmas}\} or 2 templates: \texttt{Las CITY1 is good.}, \texttt{CITY1=}\{\textit{Vegas, Palmas}\} and \texttt{New CITY2 is good.}, \texttt{CITY2=}\{\textit{York, Delhi}\}. This problem is hard to solve without heuristics. One possibility is to use the translation alignment information however, such alignments are often imperfect even for high-resource languages. We leave improvements to TEA for handling this to future work. 
 
\paragraph{Handling Morphology} Creating good templates for morphologically rich languages \cite{sinha-etal-2005-translation,dorr-1994-machine} is more challenging due to inflections. For e.g, in Hindi a verb may take different form for different tenses and gender. While TEA can handle such cases by creating multiple templates, 
but with still a third of Hi templates needed correcting. We leave morphologically informed CheckList creation to future work.

 \paragraph{Translation Errors} Translation errors are a frequent pattern, affecting the input target language instances. 
 In some cases, due to the statistical nature of TEA, we are able to naturally filter out such erroneous templates. 
 For e.g, for an En template 'I used to think this \{air\_noun\} was \{neg\_adj\}, \{change\} now I think it is \{pos\_adj\}', translated Hi templates 'Mujhe lagta hai ki us \{udaan\} \{ghatia\} tha, ab mujhe lagta hai ki yeh {asadharan} hai' (correct) matches 187 translations, and 'Mujhe lagta hai ki us \{udaan\} \{ghatia\} tha \textcolor{red}{karte the}, ab mujhe lagta hai ki yeh bohut achha hai' (noisy) matches only 35. While TEA can remove some noisy patterns, errors due to misunderstood context are much harder to fix. For e.g 'the service is poor' translated as 'vah seva \textcolor{red}{garib} hai' but `garib' in Hindi means ``lacking sufficient money" and {\em not} ``lower or insufficient standards". We leave comparisons of TEA for human v.s machine translated input instances and methods to measure and reduce the effect of translation errors on TEA to future work.
 
\paragraph{Metric Limitations} Quantifying  the quality of generated template and verification of the relevance of templates with respect to provided description is non-trivial . While we suggest a set of metrics quantifying utility, diversity and cost, these should be extended and further studied for efficacy across tasks and languages. Lastly, soundness and completeness of a template sets (or a test-suite in general) is another unexplored aspect in our current work and an important future direction of research. Furthermore, we acknowledge the limitation of Failure Rate as a metric in the sense that the model could also fail if an example is ungrammatical. In other words, FR is conditional to correctness of the CheckList. However, in our experimentation in both Hindi and other languages, we have found that the the difference between the FR of human verified TEA-ver and TEA is typically small (with a few exceptions) across languages. This means that high FR being caused due to ungrammatical instances here is unlikely. Thus, as stated before, the closeness of the FRs of TEA and TEA-ver points to the reliability of the TEA algorithm.
 

%% file: aacl/9_conclusion.tex
In this paper we proposed TEA (\textbf{T}emplate \textbf{E}xtraction \textbf{A}lgorithm) to automatically generate multilingual CheckLists in a target language without any human supervision (\S\ref{3_tea}). This algorithm recursively extracts templates and lexicon from an input set of instances by treating sentences as a directed acyclic graph of words and combining them.

We additionally experimented with 3 other approaches with varying degrees of human intervention, 2 manual and 1 semi-automatic for CheckList generation (\S\ref{4_multilingual_checklist_generation}). For comparing these CheckLists, we introduced metrics along the dimensions of utility, diversity, cost and correctness (\S\ref{5_metrics}). 

We performed in-depth analysis of all the 4 methods, with varying degree of human interventions, to create CheckLists for Sentiment Analysis and NLI in Hindi (\S\ref{6_hindi_checklists}). In addition to Hindi, we experimented with 9 more typologically diverse languages to demonstrate the efficacy of TEA along with comparison with human-verified CheckLists in 3 of them (\S\ref{4_multilingual_checklist_generation}). We found that TEA is cost-effective, useful, and diverse in the CheckLists that it generates. While around one-third of the TEA templates required correction by humans, making the semi-automatic approach more reliable, we find that the model performance estimates provided by unverified CheckLists are very close to that of the human-verified (or semi-automatically created) CheckLists and are also significantly correlated to it. We also substantiated the finding of TEA being effective as well as reliable in the other languages. 

Our overall recommendation is that TEA followed by human verification is the most reliable and cost-effective way to scale CheckList evaluation to multiple languages. But in case of very limited resources, TEA is still good enough to test system performance. We end with a discussion on the limitations of this work and propose directions that will, hopefully, inspire research in scaling and improving multilingual evaluation using CheckLists. Finally, we note that TEA is general purpose algorithm of template extraction that can be used for other template-based evaluations such as bias evaluation \cite{webster2020measuring, bhatt2022re}

%% file: aacl/app_1_tea_details.tex
\section{Details of TEA}
\label{app:tea_details}

\subsection{Template as a Grammar}
A template can be considered as a type of grammar to generate sentences. Consider the template T0 introduced below.
\begin{quote}
    T0: \texttt{CITY-0} is beautiful but \texttt{CITY-1} is bigger.\\
    \texttt{CITY} =  \{Delhi, Paris, New York\} ,
\end{quote}
Here, the keywords (\texttt{CITY-0},\texttt{CITY-1}) are the non-terminals and their corresponding lexicons are the terminal symbols. Also, \texttt{CITY-1} should be different than \texttt{CITY-0}; and hence the non-terminal symbols cannot be replaced independently of each other, establishing the context-sensitive nature of templates. This is a why we need to look beyond probabilistic context free grammar induction to learn the templates. 

\noindent \textbf{Convention and Assumptions:} We use {\em terminal} and {\em non-terminal} to denote {\em lexicons} and {\em keywords} respectively. In a template, if the non-terminals are appended with cardinals from 0 to $k$, then they can \em{not} be replaced with same terminal while generating sentences. Also, if a template contains an instance of a non-terminal with cardinal $k, (k>0)$ then at least one instance of the same non-terminal with cardinal $k-1$ should have occurred before its occurrence in the template.

\subsection{TEA Algorithm}
We first briefly recap the pipeline of TEA for ease of exposition.  We start with an En template and corresponding terminals created by a human expert, and generate a set of examples by substituting the non-terminals with their appropriate terminals. We then translate the examples to Hi using an Automatic Machine Translation system (such as Azure cloud Translator). Then we extract Hi template(s), terminal(s) and non-terminal(s) from the Hi examples. The process of extracting Hi templates are repeated for each of the En templates, providing us a (tentative) CheckList for Hi. 
Here, we describe in detail the TEA algorithm that extracts Hi templates (along with Hi terminal words) from the Hi examples. First we discuss our approach to extract potential set of terminal words, i.e., we group a set of words (terminals) and give them a symbol/name (non-terminal). Then we extract the templates using the terminals and non-terminals that are extracted in previous step. Towards the end of this section, we briefly discuss the scalability issues and the approximations that we used to make it more scalable. 

\subsubsection{Extracting and Grouping Terminals}\label{sec:find_lexicon}

First, we convert the given Hi examples into a directed graph whose nodes are unique words (or tokens, if we use a different tokenizer) from the examples and there is an edge from word A to word B if word B follows word A in at least one of the examples. In this directed graph (as shown in Fig.~\ref{fig:tea}), between any two nodes, if there are multiple paths of length less than equal to $k+1$, we group all those paths and give the group a name or a non-terminal symbol (for example Key\_1 and Key\_2 in Fig.~\ref{fig:tea}).\footnote{We assumed the maximum length of each terminal string to be $k (=2)$ tokens/words} By grouping the paths, we meant to concatenate the intermediate words in the path (with space in between them) and then to group the concatenated strings (terminals). This step gives us potential lexicons and keywords (or list of terminals grouped together). 

\subsubsection{Template Extraction given Terminal and Non-Terminals}\label{sec:extract_template}

Input to our algorithm is (1) a set Hi examples denoted by $S = \{s_1, s_2, \ldots s_N \}$, and (2) all terminals (denoted by $w$) and its corresponding non-terminals (denoted by $v$) that are extracted in previous step  $\forall i, v_i = {w_{i1}, w_{i2}, \ldots}$ In other words, these are the production rules from a non-terminal to (only) terminals. Output of our algorithm is a set of templates $\hat{T} = \{t_1, t_2, ...\} $ such that $\hat{T}$ can generate all the examples in $S$ using only the given non-terminal and their corresponding terminals. 

For convenience, we represent non-terminals and its corresponding terminals as a list (or ordered set) of $\langle$ terminal, non-terminal $\rangle$ tuples, the list is denoted by $L = [\langle w_1, v_1 \rangle... \langle w_i, v_i \rangle ... ] $. The tuple $\langle w_i, v_i \rangle$ belongs to $L$ if and only if the the terminal $w_i$ belongs to the non-terminal $v_i$. 

The trivial result for $\hat{T}$ is $S$ itself, as $S$ can generate every example (using no terminals). But this is not useful because, the essence of extracting templates from a set of examples is that one should be able to read/write the entire set by reading only a few templates. Therefore, the objective is to find the (approximately) smallest $\hat{T}$ such that it can generate entire $S$. 

\setlength{\textfloatsep}{0pt}
\begin{algorithm}[t]
\caption{Extract templates given terminals and non-terminals}\label{alg:extract_template}
\begin{algorithmic}[1]
\Require{$S = \{s_1, s_2, ....s_N \}$,  $L = [\langle w_1, v_1 \rangle... \langle w_i, v_i \rangle ... ]$ }
\Ensure{$\hat{T}$, the approximately smallest set of templates that generates entire $S$}
\For{each $s_i$ in S}
    \State $T_i \gets $ \Call{Get-Templates-Per-Example}{$s_i, L$}
\EndFor
\State Find (approximately) smallest $\hat{T}$ such that $\forall T_i,  \hat{T} \cap T_i \neq \emptyset$ \Comment{Variant of set cover, use greedy approach}
\State \textbf{return $\hat{T}$}

\Procedure{Get-Templates-Per-Example}{$s_i, L$}
\State $T_i  \gets \{s_i\}$
\For{each $\langle w_m, v_m \rangle$ in $L$}
    \State $T_{new} \gets \{\}$
    \For{each $t_{ij}$ in $T_i$ }
        \If{$w_m$ is sub-string of $t_{ij}$}
            \State $t_{new} \gets $ \Call{Replace-Matched-String}{$t_{ij}, w_m, v_m$} \Comment{Refer \S \ref{sec:Replace-Matched-String}}
            \State $t_{new} \gets $ \Call{Rename-Nonterminal-cardinals}{$t_{new}$} \Comment{Refer \S \ref{sec:Rename-Nonterminal-cardinals}}
            \State $T_{new} \gets T_{new} \cup t_{new} $
        \EndIf
    \EndFor 
    \State $T_i  \gets T_i \cup T_{new}$ 
\EndFor
\State \textbf{return} $T_i$
\EndProcedure
\end{algorithmic}
\end{algorithm}

We provide the outline of our algorithm in Algorithm~\ref{alg:extract_template}. Next, we explain the algorithm along with the helper functions that are not elaborated in the pseudocode. 
For each sentences $s_i$, we call the function \textsc{Get-Templates-Per-Example} to generate a set of templates, $T_i = \{t_{i1}, t_{i2}, ...\}$, such that $s_i$ belongs to the set of examples generated by each $t_{ij}$. 
Once we have $T_i$ for every $s_i$, we construct the (approximately) smallest set $\hat{T}$ such that $\forall i, \hat{T} \bigcap T_i \neq \emptyset $. Note that for every sentence $s_i \in S$, there exist atleast one template in $\hat{T}$ that generates $s_i$. Finding the smallest $\hat{T}$ is a variant of set cover problem, therefore we use greedy approach to find the approximately small $\hat{T}$. 






\noindent \textbf{Generating $T_i$:} For every terminal string ($w_m$) that is a substring of example $s_i$ (or intermediate template $t_i$), we have 2 options to create template, either (1) replace the matched substring ($w_m$) with its corresponding non-terminal ($v_m$) or (2) leave as it is; we can make this decision to replace or not, independently for every matched terminals. While replacing, we need to take care of the cardinals for non-terminals and make sure the templates conform to the adopted convention. 
We use the functions \textsc{Replace-Matched-String} and \textsc{Rename-Nonterminal-cardinals} to ensure such conformance. 

\paragraph{\textsc{Replace-Matched-String}} \label{sec:Replace-Matched-String}
This function replaces the matched terminal $w_{m}$ in $t_{ij}$ with its corresponding non-terminal $v_{m}$. If there are multiple $w_{m}$ in $t_{ij}$, then each $w_{m}$ will be independently replaced with $v_{m}$ or left unchanged. For example, consider the initial template and $\langle$ terminal, non-terminal $\rangle$ pair be "\#Paris is beautiful. \texttt{CITY-0} is cold. Paris is bigger." and $\langle$ Paris, \texttt{CITY} $\rangle$ respectively. This will generate 3 templates after replacement. 
(1) "\#\texttt{CITY-1} is beautiful. \texttt{CITY-0} is cold. Paris is bigger." 
(2) "\#Paris is beautiful. \texttt{CITY-0} is cold. \texttt{CITY-1} is bigger." 
(3) "\#\texttt{CITY-1} is beautiful. \texttt{CITY-0} is cold. \texttt{CITY-1} is bigger." 

Note that, we do not search if the words in the $s_i$ is a terminal, rather we search if the terminal is a sub-string of $s_i$ (or $t_{ij}$). This makes it possible for the terminal to be a sub-word or a multi-word string and still match. Sub-word level match can be quite useful, especially in morphologically rich languages; using only the base word as lexicons it may be possible to match different morphological forms. 

\paragraph{\textsc{Rename-Nonterminal-cardinals}} \label{sec:Rename-Nonterminal-cardinals}
This function renames the cardinals to make sure that an instance of a non-terminal with cardinal $k-1$ occurs before the instance of that non-terminal with cardinal $k, (k>0)$. For example, after re-naming the cardinals, the above three templates become the following three, respectively. 
(1) "\#\texttt{CITY-0} is beautiful. \texttt{CITY-1} is cold. Paris is bigger." 
(2) "\#Paris is beautiful. \texttt{CITY-0} is cold. \texttt{CITY-1} is bigger." 
(3) "\#\texttt{CITY-0} is beautiful. \texttt{CITY-1} is cold. \texttt{CITY-0} is bigger."

\subsubsection{Combine both the steps}

First, we find all the potential terminals and non-terminals (using 
\S~\ref{sec:find_lexicon}) for all Hi examples, and then use them to extract template following the algorithm outlined in \S~\ref{sec:extract_template}. While this simple procedure is possible, it is often computationally expensive; one of the reasons is that due to noise (many of the translated sentences may not fit into a template), the algorithm to extract terminals and non-terminals (\S~\ref{sec:find_lexicon}) often gives a lot of different non-terminals that share many common terminals. For example, we may get two non-terminals with their corresponding terminals such as ``\{Paris, New York, Delhi\}''  and  ``\{London, New York, Delhi\}''. Moreover, the complexity of the algorithm in \S~\ref{alg:extract_template} to extract templates can be increased exponentially with the number of non-terminals. To mitigate this problem, we follow an iterative approach where instead of using all the extracted non-terminals (along with their terminals), we initialize the set of non-terminals with an empty set and iteratively add the most useful non-terminals (with their corresponding terminals) to the existing set of non-terminals.


%% file: aacl/app_2_checklist_capabilities_tested.tex
\section{Capabilities tested using CheckList}
\label{app:capabilities}
Capabilities are tested using MFTs. MFTs (Minimum Functionality Tests) are tests similar to unit tests in software testing where a specific pointed capability of a model is tested via a template and an expected label(s). The test is said to pass for an instance if the model predicted label matches the expected label(s). Finally, failure rate is recorded as the \% of test instances that fails; which can also be inferred as 100-accuracy. 

\subsection{Sentiment Analysis (SA)}
These capabilities, their descriptions, examples and their original template sets used in testing are all sourced from \citet{checklist-paper}.

\paragraph{Vocabulary} This capability tests whether the model can appropriately handle the impact of words with different parts of speech on the task. In particular, sentences with  neutral adjectives are expected to have a neutral prediction and sentences sentiment-laden (positive or negative) adjectives are expected to have the corresponding label. For example, ``This is a {\em private (NEUTRAL\_ADJ)} aircraft'' should be labelled neutral; and ``This is a {\em great (POSITIVE\_ADJ)} aircraft''  ``This is a {\em bad (NEGATIVE\_ADJ)} aircraft'' should be labelled positive and negative respectively. 
\paragraph{Negation} This capability tests that the negation of a positive adjective in the sentence should be labelled as positive or neutral, for example: ``This is {\em not} a {\em great (POSITIVE\_ADJ)} aircraft'' should be labelled negative or neutral. Similarly, sentence with negation of negative adjective should be positive our neutral and those with negation of neutral adjectives should remain neutral.

\paragraph{Semantic Role Labeling (SRL)} SRL aims to test that the model understands the agent, object etc in an instance. That is sentiment of the correct role in the instance is parsed. Here, there are two distinct capabilities MFTs. The first one is to test that the sentiment author sentiment is given more importance than of sentiment of others. For example, ``Some people think this aircraft is bad, but I thought it was {\em great (POSITIVE\_ADJ)}'' should be labelled as Positive. The second test is related to parsing yes/no questions with the correct sentiment. For example, ``Do I think this aircraft is great? {\em Yes}'' should be labelled as positive, whereas if the answer was No, it should be negative.

\paragraph{Temporal} This capability is used to test whether the model understands the sequence of events correctly. In other words that the most recent sentiment is correctly parsed in labelling. For example, ``I used to hate this aircraft, but {\em now I love it}'' should be labelled positive.

\paragraph{Robustness} There are two tests for robustness: First changing of values within semantically equivalent classes should not change the prediction. For example, ``I flew in from {\em Delhi}'' and ``I flew in from {\em New York}'' should have the same label as the change here is within the semantically equivalent class of 'CITY'. Secondly, typos (or random character exchange) should not flip labels. For example, ``This is a {\em graet} aircraft'' should still remain positive.

\paragraph{Fairness} Fairness is used to test that prediction should be the same for various adjectives within a protected class. For example, ``Mary is a {\em black (RACE)} woman'' and ``Mary is a {\em white (RACE)} woman'' should have same sentiment prediction.

\subsection{Natural Language Inference (NLI)}
We use the template sets from \citet{nlichecklist} which in turn rely on the taxonomy of capabilities from \cite{joshi2020taxinli} for their selection of capabilities. In examples that follow, P stands for Premise and H for hypothesis.

\paragraph{Co-reference resolution} Test the model for resolving pronouns between the premise and hypothesis correctly. For example, P: Angelique and Ricardo are colleagues. He is a minister and she is a model. H: Angelique is a model. Here H should `entail' P.

\paragraph{Spatial reasoning} Tests the model for reasoning using spatial properties. For example, P: Manchester is 67 miles from Pittsburg and 27 miles from Kansas. H: Manchester is nearer to Kansas than Pittsburg. Here H should `entail' P.

\paragraph{Causal reasoning} Tests the model for using causation in the premise to infer the hypothesis. For example, P: Katherine taught science to Nancy. H: Nancy learnt science from Katherine. Here H should `entail' P.

\paragraph{Conditional reasoning} Tests the model for logically inferring the hypothesis given conditional premise. For example, P: If the baby is fed on time, he does not get cranky. H: The baby gets crancky when he is hungry. Here H should `entail' P.

\paragraph{Comparative reasoning} Tests models for reasoning involving comparisons of objects. For example, P: The earth is larger than the moon but smaller than sun. H: The moon is smaller than sun. Here H should `entail' P.